\documentclass[conference]{IEEEtran}

\usepackage[numbers]{natbib}
\usepackage{multicol}
\usepackage{multirow}
\usepackage[colorlinks,citecolor=blue, bookmarks=true]{hyperref}

\usepackage[utf8]{inputenc} 
\usepackage[T1]{fontenc}    
\usepackage{url}            
\usepackage{booktabs}       
\usepackage{amsfonts}       
\usepackage{nicefrac}       
\usepackage{microtype}      
\usepackage{amsmath,amssymb}
\usepackage{mathtools}
\usepackage{array}

\usepackage{algorithm}
\usepackage[noend]{algpseudocode}
\usepackage{subcaption}
\usepackage[table]{xcolor}
\usepackage{adjustbox}
\usepackage{wrapfig}
\usepackage{mdframed}
\usepackage{makecell}
\usepackage{siunitx}
\usepackage{flushend}
\usepackage{todonotes}

\makeatletter
\newcommand{\vast}{\bBigg@{4}}
\newcommand{\Vast}{\bBigg@{5}}
\makeatother

\makeatletter
\def\hiddenfootnote{\gdef\@thefnmark{}\@footnotetext}
\makeatother

\newcolumntype{L}[1]{>{\raggedright\arraybackslash}p{#1}}

\title{Probabilistic Multimodal Modeling for Human-Robot Interaction Tasks}

\author{Joseph Campbell, Simon Stepputtis, and Heni Ben Amor\\
School of Computing, Informatics, and Decision Systems Engineering, Arizona State University\\
{\tt\small\{jacampb1, sstepput, hbenamor\}@asu.edu}}

\begin{document}

\maketitle

\begin{abstract}
Human-robot interaction benefits greatly from multimodal sensor inputs as they enable increased robustness and generalization accuracy.
Despite this observation, few HRI methods are capable of efficiently performing inference for multimodal systems.
In this work, we introduce a reformulation of Interaction Primitives which allows for learning from demonstration of interaction tasks, while also gracefully handling nonlinearities inherent to multimodal inference in such scenarios. We also empirically show that our method results in more accurate, more robust, and faster inference than standard Interaction Primitives and other common methods in challenging HRI scenarios.
\end{abstract}

\IEEEpeerreviewmaketitle

\section{Introduction}
\label{sec:introduction}
Human-robot interaction (HRI) requires constant monitoring of human behavior in conjunction with proactive generation of appropriate robot responses. This decision-making process must often contend with high levels of uncertainty due to partial observability, noisy sensor measurements, visual occlusions, ambiguous human intentions, and a number of other factors. The inclusion of sensor measurements from a variety of separate modalities, e.g., cameras, inertial measurement units, and force sensors, may provide complementary pieces of information regarding the actions and intentions of a human partner, while also increasing the robustness and safety of the interaction. Even in situations in which a complete sensor modality becomes temporarily unavailable, i.e., due to a hardware failure, other available modalities may ensure graceful degradation of the system behavior. Hence, it is critical to support decision-making in HRI with inference and control methods that can deal with a variable number of data sources, each of which may have distinctive numerical and statistical characteristics and limitations.
\hiddenfootnote{Library source code and video available at:\\ 
\hspace*{6pt} \href{http://interactive-robotics.engineering.asu.edu/interaction-primitives}{http://interactive-robotics.engineering.asu.edu/interaction-primitives}}

\begin{figure}[t!]
\centering
\includegraphics[width=0.95\columnwidth]{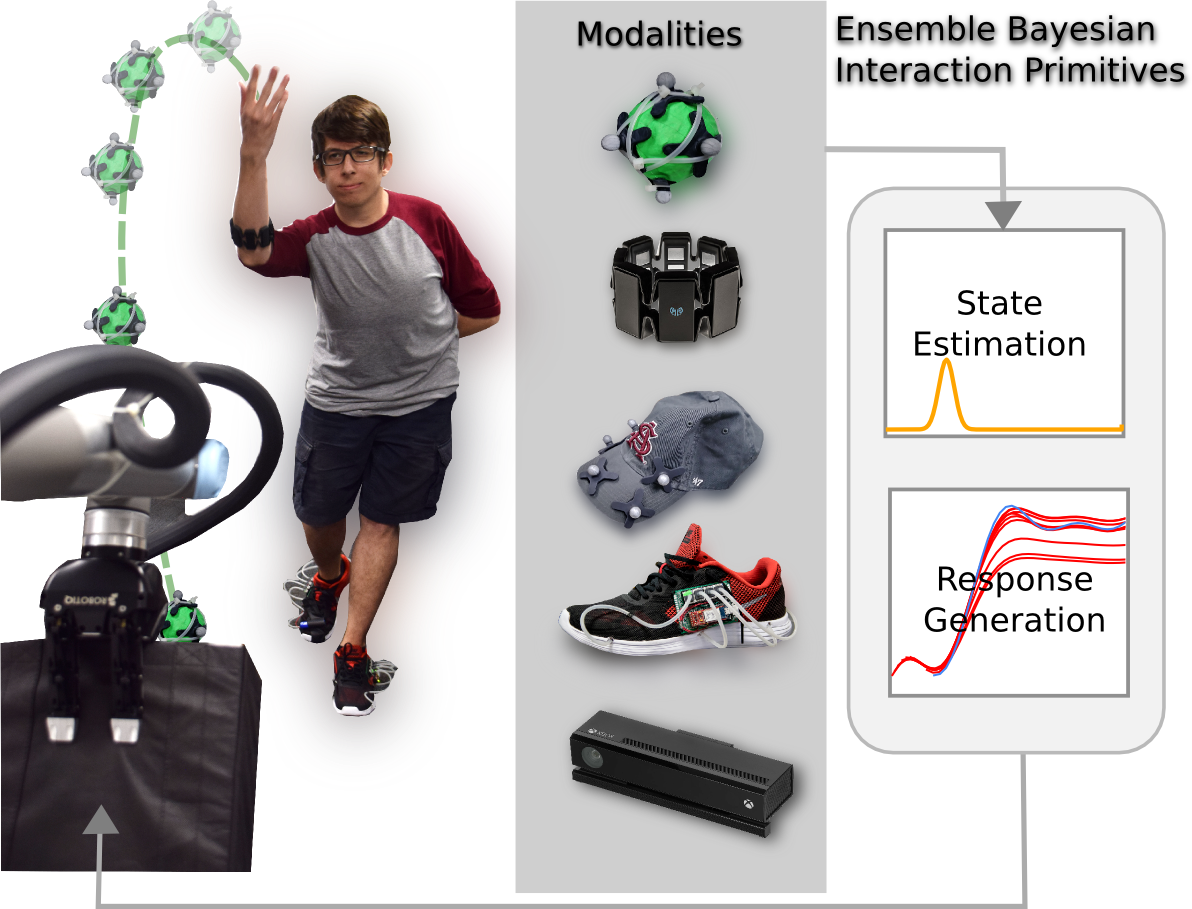}
\caption{A robot learning to catch a thrown ball by combining information from different modalities.}
\label{fig:teaser}
\end{figure}

In this paper, we investigate how multimodal models of human-robot interaction can be efficiently learned from demonstrations and, later, used to perform reasoning, inference, and control from a collection of data sources. Fig.~\ref{fig:teaser} depicts a motivating example -- a robot arm catching a ball. In this example, the position of the ball can be continuously tracked using motion capture markers, but is occluded from view while in the human's hand.
Yet, even before the ball is released from the hand, the robot may already intuit the moment of release and travel direction by reading pressure information from a smart shoe, from inertial measurements on the throwing arm, or from human pose data acquired via skeletal tracking.
Integrating these pieces of information together, we would expect the robot to generate earlier and better predictions of the ball, as well as better estimates of necessary control signals to intercept it.

Few probabilistic inference methods for HRI have examined reasoning across multiple modalities as in the above example, with many instead opting to construct models relating only two modalities, e.g., a single observed modality to a single controlled modality. In the case of Bayesian Interaction Primitives (BIP)~\cite{campbell2017bayesian}, demonstrations of two (human) agents interacting with each other are used to form a joint probability distribution among all degrees of freedom (DoF) and all modalities. During inference, this distribution is used as the prior for Bayesian filtering, which is then refined through sensor observations of the observed modalities and subsequently used to infer the controlled DoFs. However, when multiple sensing modalities are employed, several challenges quickly arise. First, in order to maintain computational tractability of the filtering process, limiting assumptions are made both about the form of the joint probability distribution, i.e., unimodal and Gaussian, as well as the linearity of the system as a whole. As the number of sensing modalities increases, each with their own unique statistical characteristics, these assumptions begin to negatively impact inference accuracy. Second, expanding the sensing modalities translates to an increased number of degrees of freedom which magnifies the computational burden and jeopardizes real-time performance -- a vital property in HRI contexts.

In this work, we propose ensemble Bayesian Interaction Primitives (eBIP) for human-robot interaction scenarios. In particular, the following contributions will be made:

\begin{enumerate}
\item An alternative formulation of interaction primitives that is particularly well-suited for inference and control in the presence of many input modalities, as well as noisy and missing observations. 
\item An ensemble-based approach to Bayesian inference for HRI, which combines advantages of parametric and non-parametric methods. The approach requires neither an explicit covariance matrix, nor a measurement model. Measurement errors are efficiently calculated in closed-form. 
\item Our approach allows for inference in nonlinear interactive systems, while avoiding typical inaccuracies due to either linearization errors, the parametric family of the prior, or the underlying dimensionality. Training demonstrations are used to model the non-Gaussian prior distribution of a task. The non-parametric nature of this prior avoids computational overheads and inaccuracies as found, for example, when fitting a mixture model.  
\item Fast and efficient inference that scales particularly well with increasing dimensionality of the task. 
\end{enumerate}

We compare eBIP to other methods on a fast-paced, dynamic human-robot interaction experiment involving multimodal sensor streams. Experiments show that eBIP allows for accurate and rapid inference in high-dimensional spaces.

\section{Related Work}
\label{sec:related_work}
In the following section, we will review relevant work on probabilistic modeling of joint actions and multimodal modeling. For a detailed discussion of computational techniques in the HRI domain, see the excellent surveys in~\cite{iqbal2019human, thomaz2016computational}.

\begin{figure*}[tp]
\centering
\includegraphics[width=0.75\textwidth]{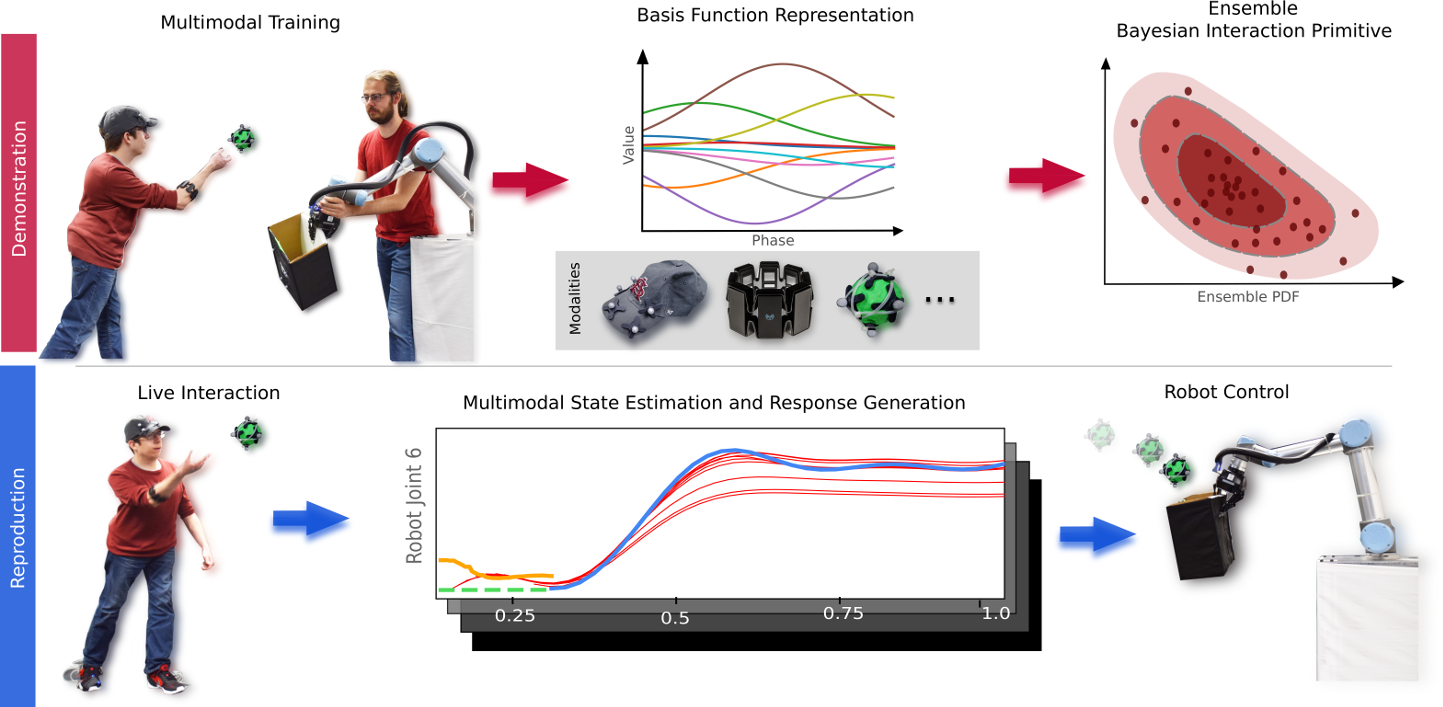}
\caption{An overview of eBIP. Top: training demonstrations (left) are decomposed into a latent space (middle) and transformed into an ensemble of samples (right). Bottom: observations are collected during a live interaction (left) which is used to perform filtering with the learned ensemble (middle) and produce a response trajectory (right).}
\label{fig:overview}
\end{figure*}

\noindent
\textbf{Probabilistic Modeling of Joint Actions: }
Early work on modeling HRI scenarios using probabilistic representations focused on HMMs~\cite{rabiner1989} as a method of choice, see for instance the works in~\cite{donghuilee2010, Rozo16Frontiers}. The ability of HMMs to perform inference in both time and space, makes them particularly interesting for collaborative and interactive tasks. However, these advantages come at a cost -- HMMs require a discretization of the state space and do not scale well in high-dimensional spaces. The concept of Interaction Primitives (IP) was first proposed in~\cite{amor2014interaction} as an alternative approach for \emph{learning from demonstration}. Intuitively, an IP models the actions of one or more agents as time dependent trajectories for each measured degree of freedom. The approach has gained popularity in HRI and has been applied to a number of tasks ~\cite{Cui2018, campbell2017bayesian, ogur2018, dermy2019multi,chen2017, ewerton2015}.

Most recently, in ~\cite{campbell2017bayesian} a fully Bayesian reformulation of IPs called \emph{Bayesian Interaction Primitives} (BIP) was introduced. Most importantly, this work establishes a theoretical link between HRI and joint optimization frameworks as found in the Simultaneous Localization and Mapping (SLAM) literature~\cite{thrun2005probabilistic}. The resulting inference framework for BIPs was shown to produce superior space-time inference performance when compared to previous IP formulations. 

\noindent
\textbf{Multimodal Modeling and Inference: }
Multimodal integration, inference and reasoning has been a longstanding and challenging problem of artificial intelligence~\cite{Coen:2001} and signal processing~\cite{Lundquist:2015}. Following the principles formulated by Piaget~\cite{piaget1955child}, many existing multimodal systems separately process incoming data streams of different types, deferring the integration step to later stages of the processing pipeline. In~\cite{Calinon08ICRA}, Calinon and colleagues present a fully probabilistic framework in which social cues from the gaze direction and speech patterns of a human partner are incorporated into the robot movement generation process. The multimodal inference process is achieved by modeling such social cues as prior probability distributions. In a similar vein, the work by Dermy et al.~\cite{dermy2019multi} uses joint probability distributions over human-robot joint actions, in order to infer robot responses to human visual or physical guidance. However, the approach assumes a fixed user position at training and test time and models the phase variable according to a predetermined relationship to the execution speed. More recently, deep learning approaches for multimodal representations have gained considerable attention. A prominent methodology is to process each data modality with separate sub-networks, which are integrated at a shared final layer~\cite{noda2014multimodal, rao2017multimodal}.
However, such neural network approaches are not well-suited for probabilistic data integration and do not provide an estimate of the uncertainty inherent to observations or outputs.
Also, such approaches cannot cope with missing inputs or changing query types, i.e., any change to the number or type of inputs requires a complete retraining of the network.

\section{Preliminaries: Bayesian Interaction Primitives}
\label{sec:prelim}

The concept of Bayesian Interaction Primitives~\cite{campbell2017bayesian} refers to a human-robot interaction framework which focuses on extracting a model of the interaction dynamics as found in example demonstrations. Given training demonstrations of the interaction task, e.g., a set of human-human interactions, BIPs can be used to capture the observed relationships between the interacting agents. Fig.~\ref{fig:overview} depicts the training and reproduction process in the BIP framework. After collecting examples for throwing and catching, the training data is represented within a basis function space and encoded as a prior distribution. The distribution is then used during reproduction to perform Bayesian filtering of live human movements, thereby enabling (a) the prediction of the next human movements, and (b) the generation of appropriate robot actions and responses. The basic structure of the above figure applies to both the original BIP formulation and our proposed method. Implementational details, in particular regarding the encoding of multiple modalities, the representation of the joint distribution as an ensemble, as well as the multimodal filtering process differ substantially in our reformulation. Subsequently, we will first provide a discussion of the BIP method as originally proposed in~\cite{campbell2017bayesian}. In particular, we will discuss the basis function decomposition and the Bayesian filtering process in BIP. After that, we introduce our main contribution called ensemble Bayesian Interaction Primitives in Sec.~\ref{sec:methodology}.   

Notation: we define an interaction $\boldsymbol{Y}$ as a time series of $D$-dimensional sensor observations over time, $\boldsymbol{Y}_{1:T} = [\boldsymbol{y}_1, \dots, \boldsymbol{y}_T] \in \mathbb{R}^{D \times T}$. Of the $D$ dimensions, $D_o$ of them represent \emph{observed} DoFs from one agent (the human) and $D_c$ of them represent the \emph{controlled} DoFs from the other agent (the robot), such that $D = D_c + D_o$.

\subsection{Basis Function Decomposition}
\label{sec:prelim_basis_decomp}

Working with the time series directly is impractical due to the fact that the state space dimension would be proportional to the number of observations, so we transform the interaction $\boldsymbol{Y}$ into a latent space via basis function decomposition.
Each dimension $d \in D$ of $\boldsymbol{Y}$ is approximated with a weighted linear combination of time-dependent basis functions: $[y^d_{1}, \dots, y^d_{t}] = [\Phi_{\phi(1)}^{d} \boldsymbol{w}^d + \epsilon_y, \dots, \Phi_{\phi(t)}^{d} \boldsymbol{w}^d + \epsilon_y]$, where $\Phi_{\phi(t)}^d \in \mathbb{R}^{1\times B^d}$ is a row vector of $B^d$ basis functions, $\boldsymbol{w}^d \in \mathbb{R}^{B^d \times 1}$, and $\epsilon_y$ is i.i.d. Gaussian noise.
As this is a linear system with a closed-form solution, the weights $\boldsymbol{w}^d$ can be found through simple linear regression, i.e., least squares.
The full latent model is composed of the aggregated weights from each dimension, $\boldsymbol{w} = [\boldsymbol{w}^{1\intercal}, \dots, \boldsymbol{w}^{D\intercal}] \in \mathbb{R}^{1 \times B}$ where $B = \sum_{d}^{D} B^d$ and we denote the basis transformation as $\boldsymbol{y}_t = h(\phi(t), \boldsymbol{w})$.

We note that the time-dependence of the basis functions is not on the absolute time $t$, but rather on a relative phase value $\phi(t)$.
Consider the basis function decompositions for a motion performed at slow speeds and fast speeds with a fixed measurement rate.
If the time-dependence is based on the absolute time $t$, then the decompositions will be different despite the motion being spatially identical.
Thus, we substitute the absolute time $t$ with a linearly interpolated relative phase value, $\phi(t)$, such that $\phi(0) = 0$ and $\phi(T) = 1$.
For notational simplicity, from here on we refer to $\phi(t)$ as simply $\phi$.

\subsection{Bayesian Filtering in Time and Space}
\label{sec:prelim_filtering}

Given $t$ observations of an interaction, $\boldsymbol{Y}_{1:t}$, the objective in BIP is to infer the underlying latent model $\boldsymbol{w}$ while taking into account a prior model $\boldsymbol{w}_0$.
We assume that the $t$ observations made so far are of a partial interaction, i.e., $\phi(t) < 1$, and that $T$ is unknown.
This requires the simultaneous estimation of the phase, as well as the phase velocity, i.e., how fast we are proceeding through the interaction, alongside the latent model. This joint estimation process is possible since the uncertainty estimates of each weight in the latent model are correlated due to a shared error in the phase estimate.
In other words, if we mis-estimate where we are in the interaction in a temporal sense, we will mis-estimate where we are in a physical sense as well.
Probabilistically, we represent this insight with the augmented state vector $\boldsymbol{s} = [\phi, \dot{\phi}, \boldsymbol{w}]$ and the following definition:
\begin{equation}
\label{eq:bip_general}
p(\boldsymbol{s}_t | \boldsymbol{Y}_{1:t}, \boldsymbol{s}_{0}) \propto p(\boldsymbol{y}_{t} | \boldsymbol{s}_t) p(\boldsymbol{s}_t | \boldsymbol{Y}_{1:t-1}, \boldsymbol{s}_{0}).
\end{equation}

The posterior density in Eq.~\ref{eq:bip_general} is computed with a recursive linear state space filter, i.e., an extended Kalman filter~\cite{thrun2005probabilistic}.
Such filters are composed of two steps performed recursively: state prediction in which the state is propagated forward in time according to the system dynamics $p(\boldsymbol{s}_t | \boldsymbol{Y}_{1:t-1}, \boldsymbol{s}_{0})$, and measurement update in which the latest sensor observation is incorporated in the predicted state $p(\boldsymbol{y}_{t} | \boldsymbol{s}_t)$.
Applying Markov assumptions, the state prediction density can be defined as:
\begin{align}
\label{eq:bip_prediction_reduced}
& p(\boldsymbol{s}_t | \boldsymbol{Y}_{1:t-1}, \boldsymbol{s}_{0}) \nonumber \\
& = \int p(\boldsymbol{s}_t | \boldsymbol{s}_{t-1}) %
p(\boldsymbol{s}_{t-1} | \boldsymbol{Y}_{1:t-1}, \boldsymbol{s}_{0})d\boldsymbol{s}_{t-1}.
\end{align}
As with all Kalman filters, we assume that all error estimates produced during recursion are normally distributed, i.e., $p(\boldsymbol{s}_t | \boldsymbol{Y}_{1:t}, \boldsymbol{s}_{0}) = \mathcal{N}(\boldsymbol{\mu}_{t|t}, \boldsymbol{\Sigma}_{t|t})$ and $p(\boldsymbol{s}_t | \boldsymbol{Y}_{1:t-1}, \boldsymbol{s}_{0}) = \mathcal{N}(\boldsymbol{\mu}_{t|t-1}, \boldsymbol{\Sigma}_{t|t-1})$.
The state evolves according to a linear constant velocity model:

\begin{align}
\boldsymbol{\mu}_{t|t-1} &= 
{\underbrace{
		\begin{bmatrix}
		1 & \Delta t & \dots & 0\\
		0 & 1 & \dots & 0\\
		\vdots & \vdots & \ddots & \vdots\\
		0 & 0 & \dots & 1
		\end{bmatrix}
	}_\text{$\boldsymbol{G}$}}
\boldsymbol{\mu}_{t-1|t-1},\\
\boldsymbol{\Sigma}_{t|t-1} &= \boldsymbol{G} \boldsymbol{\Sigma}_{t-1|t-1} \boldsymbol{G}^{\intercal} +
\underbrace{\begin{bmatrix}
	\Sigma_{\phi, \phi} & \Sigma_{\phi, \dot{\phi}} & \dots & 0\\
	\Sigma_{\dot{\phi}, \phi} & \Sigma_{\dot{\phi}, \dot{\phi}} & \dots & 0\\
	\vdots & \vdots & \ddots & \vdots\\
	0 & 0 & \dots & 1
	\end{bmatrix}}_\text{$\boldsymbol{Q}_t$},
\label{eq:ip_cov_initial}
\end{align}
where $\boldsymbol{Q}$ is the process noise associated with the state transition, e.g. discrete white noise.
The observation function $h(\cdot)$ is nonlinear with respect to $\phi$ and must be linearized via Taylor expansion:
\begin{align}
\begin{split}
\boldsymbol{H}_t &= \frac{\partial h(\boldsymbol{s}_t)}{\partial s_t}\\
&= 
\begin{bmatrix}
\frac{\partial \Phi_{\phi}^{\intercal} \boldsymbol{w}^1}{\partial \phi} & 0 & \Phi_{\phi}^1 & \dots & 0\\
\vdots & \vdots & \vdots & \ddots & \vdots\\
\frac{\partial \Phi_{\phi}^{\intercal} \boldsymbol{w}^{D}}{\partial \phi} & 0 & 0 & \dots & \Phi_{\phi}^D
\end{bmatrix}. \label{eq:bip_jacobian}
\end{split}
\end{align}
This yields the measurement update
\begin{align}
\boldsymbol{K}_t &= \boldsymbol{\Sigma}_{t|t-1} \boldsymbol{H}_t^{\intercal} (\boldsymbol{H}_t \boldsymbol{\Sigma}_{t|t-1} \boldsymbol{H}_t^{\intercal} + \boldsymbol{R}_t)^{-1},\\
\boldsymbol{\mu}_{t|t} &= \boldsymbol{\mu}_{t|t-1} + \boldsymbol{K}_t(\boldsymbol{y}_t - h(\boldsymbol{\mu}_{t|t-1})),\\
\boldsymbol{\Sigma}_{t|t} &= (I - \boldsymbol{K}_t \boldsymbol{H}_t)\boldsymbol{\Sigma}_{t|t-1}, \label{eq:bip_cov_update}
\end{align}
where $\boldsymbol{R}_t$ is the Gaussian measurement noise associated with the sensor observation $\boldsymbol{y}_t$.

The prior model $\boldsymbol{s}_0 = [\phi_0, \dot{\phi}_0, \boldsymbol{w}_0]$ is computed from a set of initial demonstrations.
That is, given the latent models for $N$ demonstrations, $\boldsymbol{W} = [\boldsymbol{w}_1^{\intercal}, \dots, \boldsymbol{w}_N^{\intercal}]$, we define $\boldsymbol{w}_0$ as simply the arithmetic mean of each DoF.
The initial phase $\phi_0$ is set to $0$ under the assumption that all interactions start from the beginning.
The initial phase velocity $\dot{\phi}_0$ is the arithmetic mean of the phase velocity of each demonstration (reciprocal length $1/T$).
The prior density is defined as $p(\boldsymbol{s}_0) = \mathcal{N}(\boldsymbol{\mu}_0, \boldsymbol{\Sigma}_0)$ where
\begin{align}
\boldsymbol{\mu}_0 &= \boldsymbol{s}_0,\label{eq:bip_prior_mean}\\
\boldsymbol{\Sigma}_0 &= %
\begin{bmatrix}
\boldsymbol{\Sigma}_{\phi, \phi} & 0\\
0 & \boldsymbol{\Sigma}_{\boldsymbol{W}, \boldsymbol{W}}
\end{bmatrix}, \label{eq:bip_prior_cov}
\end{align}
and $\boldsymbol{\Sigma}_{\phi, \phi}$ is the variance in the phases and phase velocities of the demonstrations, with no initial correlations between them.


\section{Ensemble Bayesian Interaction Primitives}
\label{sec:methodology}

The introduction of additional sensor modalities is intended to increase the robustness of the inference process for the latent model, as defined in Eq.~\ref{eq:bip_general}, by revealing additional information about the true state of the environment.
However, naively increasing the number of observed degrees of freedom often harms the inference process.
This is due to three reasons: 1) the approximation errors in the prior distribution may increase, 2) the linearization errors may increase, and 3) the state dimension increases.
This motivates the form of our proposed method as we seek to explicitly address these three issues.

\textbf{Non-Gaussian Uncertainties: }%
In general, the extended Kalman filter employed for recursive filtering in BIP relies on the assumption that uncertainty in the state prediction is approximately Gaussian.
When this is not the case, the estimated state can diverge rapidly from the true state.
One potential source of non-normality in the uncertainty is the nonlinear state transition or observation function in the dynamical system.
The original formulation of BIP addresses this challenge by linearizing these functions about the estimated state via first-order Taylor approximation, which is performed in Eq.~\ref{eq:bip_jacobian} for the nonlinear observation function $h(\cdot)$.
Unfortunately, this produces linearization errors resulting from the loss of information related to higher-order moments.
In strongly nonlinear systems this can result in poor state estimates and in the worst case cause divergence from the true state~\cite{miller1994advanced}.

As we add additional degrees of freedom from modalities with their own unique numerical fingerprint (we do not make assumptions about statistical independence, however), we potentially increase the nonlinearity of the observation model.
We follow an ensemble-based filtering methodology~\cite{evensen2003ensemble} which avoids the Taylor series approximation and hence the associated linearization errors. Fundamentally, we approximate the state prediction with a Monte Carlo approximation where the sample mean of the ensemble models the mean $\boldsymbol{\mu}$ and the sample covariance models the covariance $\boldsymbol{\Sigma}$.
Thus, rather than calculating these values explicitly during state prediction at time $t$ as in Eq.~\ref{eq:bip_cov_update}, we instead start with an ensemble of $E$ members sampled from the prior distribution $\mathcal{N}(\boldsymbol{\mu}_{t-1|t-1}, \boldsymbol{\Sigma}_{t-1|t-1})$ such that $\boldsymbol{X}_{t-1|t-1} = [\boldsymbol{x}^1,\dots,\boldsymbol{x}^E]$.
Each member is propagated forward in time using the state evolution model with an additional perturbation sampled from the process noise,
\begin{align}
\boldsymbol{x}^j_{t|t-1} &=
\boldsymbol{G} \boldsymbol{x}^j_{t-1|t-1}
+
\mathcal{N}
\left(0, \boldsymbol{Q}_t\right), \quad 1 \leq j \leq E.
\label{eq:state_prediction}
\end{align}
As $E$ approaches infinity, the ensemble effectively models the full covariance calculated in Eq.~\ref{eq:ip_cov_initial}~\cite{evensen2003ensemble}.
We note that in BIP the state transition function is linear, however, when this is not the case the nonlinear function $g(\cdot)$ is used directly.

During the measurement update step, we calculate the innovation covariance $\boldsymbol{S}$ and the Kalman gain $\boldsymbol{K}$ directly from the ensemble, with no need to specifically maintain a covariance matrix.
We begin by calculating the transformation of the ensemble to the measurement space, via the nonlinear observation function $h(\cdot)$, along with the deviation of each ensemble member from the sample mean:
\begin{align}
\boldsymbol{H}_t\boldsymbol{X}_{t|t-1} &= \left[h(\boldsymbol{x}^1_{t|t-1}), \dots, h(\boldsymbol{x}^E_{t|t-1})\right],\\
\boldsymbol{H}_t\boldsymbol{A}_t &= \boldsymbol{H}_t\boldsymbol{X}_{t|t-1} \\
&- \left[ \frac{1}{E} \sum_{j=1}^{E}h(\boldsymbol{x}^j_{t|t-1}), \dots, \frac{1}{E} \sum_{j=1}^{E}h(\boldsymbol{x}^j_{t|t-1}) \right]. \nonumber
\end{align}
The innovation covariance can now be found with
\begin{align}
\boldsymbol{S}_t &= \frac{1}{E - 1} (\boldsymbol{H}_t\boldsymbol{A}_t) (\boldsymbol{H}_t\boldsymbol{A}_t)^\intercal + \boldsymbol{R}_t,    
\end{align}
which is then used to compute the Kalman gain as
\begin{align}
\boldsymbol{A}_t &= \boldsymbol{X}_{t|t-1} - \frac{1}{E} \sum_{j=1}^{E}\boldsymbol{x}^j_{t|t-1},\\
\boldsymbol{K}_t &= \frac{1}{E - 1} \boldsymbol{A}_t (\boldsymbol{H}_t\boldsymbol{A}_t)^\intercal \boldsymbol{S}^{-1}_t.
\end{align}
With this information, the ensemble can be updated to incorporate the new measurement perturbed by stochastic noise:
\begin{align}
\boldsymbol{\tilde{y}}_t &= \left[ \boldsymbol{y}_t + \epsilon^1_y, \dots, \boldsymbol{y}_t + \epsilon^E_y \right], \nonumber \\
\boldsymbol{X}_{t|t} &= \boldsymbol{X}_{t|t-1} + \boldsymbol{K} (\boldsymbol{\tilde{y}}_{t} - \boldsymbol{H}_t\boldsymbol{X}_{t|t-1}).
\label{eq:measurement_update}
\end{align}
It has been shown that when $\epsilon_{y} \sim \mathcal{N}(0, \boldsymbol{R}_t)$, the measurements are treated as random variables and the ensemble accurately reflects the error covariance of the best state estimate~\cite{burgers1998analysis}.
The measurement noise $\boldsymbol{R}_t$ can be calculated with the following closed-form solution:
\begin{align}
    \boldsymbol{R}_t = \frac{1}{N} \sum_{i}^{N} \frac{1}{T_i} \sum_{t}^{T_i} \left( \boldsymbol{y}_{t} - h([\phi(t), \boldsymbol{w}_i]) \right)^{2}.
\end{align}
This value is equivalent to the mean squared error of the regression fit for our basis functions over every demonstration.
Intuitively, this represents the variance of the data around the regression and captures both the approximation error and the sensor noise associated with the observations.

One of the advantages of this algorithm is the elimination of linearization errors through the use of the nonlinear functions.
While this introduces non-normality into the state uncertainties, it has been shown that the stochastic noise added to the measurements pushes the updated ensemble towards normality, thereby reducing the effects of higher-order moments~\cite{lawson2004implications, lei2010comparison} and improving robustness in nonlinear scenarios.

\begin{figure}[ht]
\begin{mdframed}
\textbf{Ensemble Bayesian Interaction Primitives}

\vspace{0.7em}

\textbf{Input: } $\boldsymbol{W} = \left[ \boldsymbol{w}_1^\intercal, \dots, \boldsymbol{w}_N^\intercal \right] \in \mathbb{R}^{B \times N}$: set of $B$ basis weights corresponding to $N$ demonstrations, $\boldsymbol{l} = \left[\frac{1}{T_1}, \dots, \frac{1}{T_N]}\right] \in \mathbb{R}^{1 \times N}$: reciprocal lengths of demonstrations, $\boldsymbol{y}_{t} \in \mathbb{R}^{D \times 1}$: sensor observation at time $t$.

\vspace{0.7em}

\textbf{Output: } $\boldsymbol{\hat{y}}_{t} \in \mathbb{R}^{D \times 1}$: the inferred trajectory at time $t$.

\begin{enumerate}
    \item Create the initial ensemble $\boldsymbol{X}_0$ such that
    \begin{align}
        \boldsymbol{x}^j_0 = \left[ 0, \dot{\phi}^j, \boldsymbol{w}^j \right], \quad 1 \leq j \leq E, \nonumber
    \end{align}
    eBIP$^-$: $\boldsymbol{w}^j \sim \sum_k^K \alpha_k \mathcal{N}(\mu_k, \Sigma_k$) where $\mu_k$, $\Sigma_k$, and $\alpha_k$ are found via EM over $\boldsymbol{W}$, $\dot{\phi}^j \sim \mathcal{N}(\mu_{\boldsymbol{l}}, \sigma_{\boldsymbol{l}}^2)$ \\
    
    \vspace{0.1em}
    
    eBIP: $i \sim \mathcal{U}\{1, N\}$, $\boldsymbol{w}^j = \boldsymbol{w}_i$, $\dot{\phi}^j = \frac{1}{T_i}$.
    \vspace{0.3em}
    \item For time step $t$, propagate the ensemble forward in time as in Eq.~\ref{eq:state_prediction}:
    \begin{align}
        \boldsymbol{x}^j_{t|t-1} &=
        \boldsymbol{G} \boldsymbol{x}^j_{t-1|t-1}
        +
        \mathcal{N}
        \left(0, \boldsymbol{Q}_t\right), \quad 1 \leq j \leq E. \nonumber
    \end{align}
    \item If a measurement $\boldsymbol{y}_t$ is available, perform the measurement update step from Eq.~\ref{eq:measurement_update}:
    \begin{align}
        \boldsymbol{X}_{t|t} &= \boldsymbol{X}_{t|t-1} + \boldsymbol{K} (\boldsymbol{\tilde{y}}_{t} - \boldsymbol{H}_t\boldsymbol{X}_{t|t-1}) \nonumber
    \end{align}
    \item Extract the estimated state and uncertainty from the ensemble:
    \begin{align}
        \boldsymbol{\mu}_{t|t} &= \frac{1}{E} \sum_{j=1}^{E}\boldsymbol{x}^j_{t|t-1}, \quad
        \boldsymbol{\Sigma}_{t|t} = \frac{1}{E - 1} \boldsymbol{A}_t \boldsymbol{A}_t^{\intercal} \nonumber
    \end{align}
    \item \textbf{Output} the trajectory for each controlled DoF:
    \begin{align}
        \boldsymbol{\hat{y}}_t = h(\boldsymbol{\mu}_{t|t}) \nonumber
    \end{align}
    \item Repeat steps 2-5 until the interaction is concluded.
        
\end{enumerate}
\end{mdframed}
    \caption{Ensemble Bayesian Interaction Primitives}
    \label{fig:algorithm}
\end{figure}

\begin{figure*}[ht!]
\centering
\includegraphics[width=0.80\textwidth]{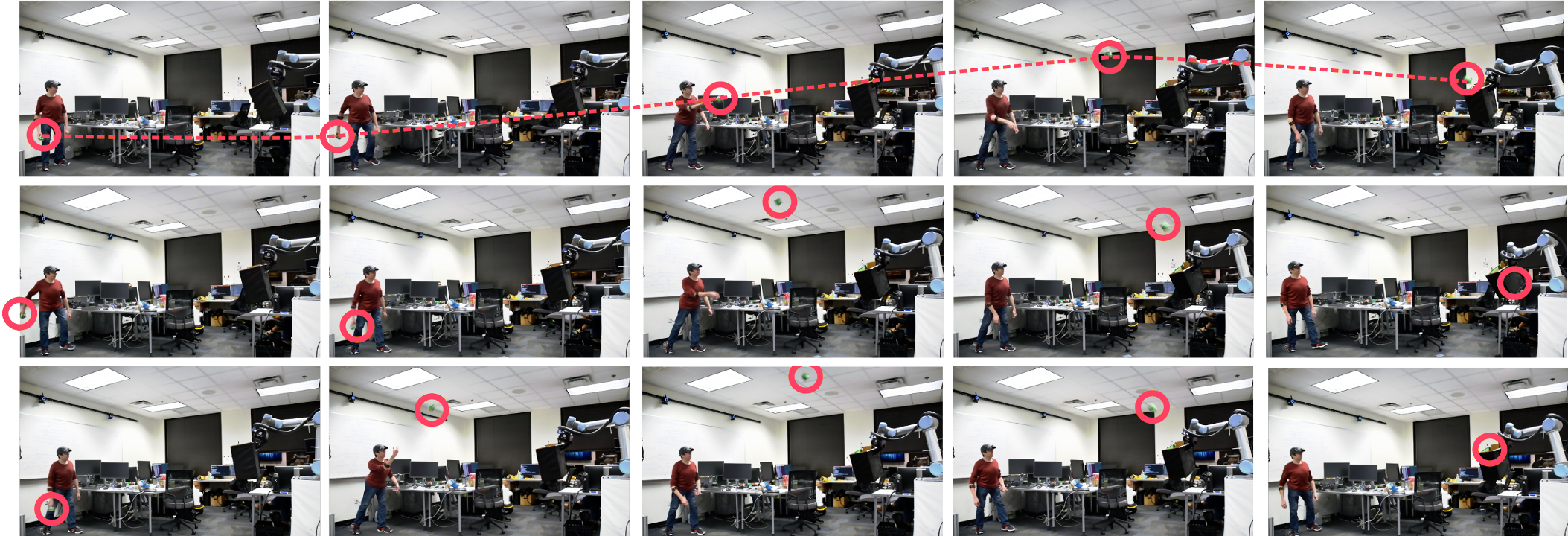}
\caption{A sequence of images from three live interactions. The robot is already reacting to the human by the second image in each sequence and catches the ball in different poses due to the different ball trajectories.}
\label{fig:throw_sequence}
\end{figure*}

\textbf{Non-Gaussian Prior: }%
Another source of non-Gaussian uncertainty is from the initial estimate (the prior) itself.
In BIP, our prior is given by a set of demonstrations which indicate where we believe a successful interaction would lie in the state space.
As we have yet to assimilate any observations of a new interaction, the (unknown) true distribution from which the demonstrations are sampled represents our best initial estimate of what it may be.
However, given that these are real-world demonstrations, they are highly unlikely to be normally distributed.
As such, two options are available in this case: we can either use the demonstrations directly as samples from the non-Gaussian prior distribution or approximate the true distribution with a Gaussian and sample from it.
The latter approach is used by BIP in Eq.~\ref{eq:bip_prior_mean} and Eq.~\ref{eq:bip_prior_cov}, however, this comes with its own risks since a poor initial estimate can lead to poor state estimates~\cite{haseltine2005critical}.
Given that the ensemble-based filtering proposed here provides a degree of robustness to non-Gaussian uncertainties, we choose to use samples from the non-Gaussian prior directly in eBIP, with the knowledge that the ensemble will be pushed towards normality.
If the number of ensemble members is greater than the number of available demonstrations, then the density of the true interaction distribution will need to be estimated given the observed demonstrations. This can be accomplished using any density estimation technique, e.g., a Gaussian mixture model, and we denote this as the alternative formulation eBIP$^-$.

\textbf{Computational Performance: }%
The increased state dimension resulting from the introduction of additional sensor modalities leads to undesirable increases in computation times in the BIP algorithm. This is due to the necessary covariance matrix updates defined in Eq.~\ref{eq:bip_cov_update}, which causes BIP to yield an asymptotic computational complexity of approximately $O(n^{3})$ with respect to the state dimension $n$~\cite{thrun2005probabilistic}; we ignore terms related to the measurement dimension as it is significantly smaller than the state dimension.
However, as eBIP is ensemble-based, we no longer explicitly maintain a covariance matrix; this information is implicitly captured by the ensemble.
As a result, the computational complexity for eBIP is approximately $O(E^2 n)$, where $E$ is the ensemble size and $n$ is the state dimension~\cite{mandel2006efficient}.
Since the ensemble size is typically much smaller than the state dimension, this results in a performance increase when compared to BIP.
Furthermore, the formulation presented in this work also obviates the need to explicitly construct the observation matrix $\boldsymbol{H}$.
The creation of the observation matrix introduces an additional overhead for BIP as it must be initialized at each time step due to the phase-dependence, a process which is unnecessary in eBIP.

In addition, we also benefit from the computational performance-accuracy trade off inherent to all sample-based methods. Inference accuracy can be sacrificed for computational performance by lowering the number of ensemble members when called for. While this is also true for particle filters, they generally scale poorly to higher state dimensions due to sample degeneracy. In particle filtering, ensemble members are re-sampled according to their weight in a scheme known as importance sampling. However, in large state spaces it is likely that only a small number of ensemble members will have high weights, thus eventually causing all members to be re-sampled from only a few. In our proposed method this is not the case, as all members are treated equally, thus lending itself well to high-dimensional state spaces.

\textbf{Algorithm: }%
Putting together all of the components, our full proposed algorithm is shown in Fig.~\ref{fig:algorithm}.

\begin{table*}[ht]
    \centering
    \begin{minipage}{.6\linewidth}
        \centering
        \begin{tabular}{ccccccc}
             \multicolumn{7}{c}{\textbf{Robot Joint Error}}\\ 
             \cmidrule(lr){2-7}\morecmidrules\cmidrule(lr){2-7}
             & & \thead{Ball} & \thead{Shoe, IMU} & \thead{Shoe, IMU\\Head} & \thead{Shoe, IMU\\Head, Ball} & \thead{All}  \\
             \cmidrule(lr){2-7}\morecmidrules\cmidrule(lr){2-7}
             \parbox[t]{2mm}{\multirow{4}{*}{\rotatebox[origin=c]{90}{43\%}}} & BIP & - & \num{3.91e15} & \num{1.48e14} & \num{9.56e14} & \num{3.97e16} \\
             & PF & - & 0.37 & 0.28 & 0.28 & 0.26 \\
             & eBIP$-$ & - & \num{5.62e6} & \num{9.98e6} & \num{1.08e7} & \num{6.00e7} \\
             & eBIP & - & \cellcolor{green!15} 0.18 & \cellcolor{green!15} 0.19 & \cellcolor{green!15} 0.18 & \cellcolor{green!15} 0.14 \\
             \cmidrule(lr){2-7}
             \parbox[t]{2mm}{\multirow{4}{*}{\rotatebox[origin=c]{90}{82\%}}} & BIP & \num{1.04e15} & \num{1.59e17} & \num{4.22e15} & \num{3.66e15} & \num{6.52e17} \\
             & PF & 0.11 & 0.37 & 0.28 & 0.28 & 0.26 \\
             & eBIP$-$ & 10.52 & \num{9.46e6} & \num{1.36e7} & \num{1.68e7} & \num{7.66e7} \\
             & eBIP & \cellcolor{green!15} 0.05 & \cellcolor{green!15} 0.20 & \cellcolor{green!15} 0.19 & \cellcolor{green!15} 0.11 & \cellcolor{green!15} 0.09 \\
             \cmidrule(lr){2-7}\morecmidrules\cmidrule(lr){2-7}
        \end{tabular}
    \end{minipage}%
    \begin{minipage}{.4\linewidth}
        \centering
        \begin{tabular}{ccccc}
            \multicolumn{5}{c}{\textbf{Ball Position Error}}\\
             \cmidrule(lr){2-5}\morecmidrules\cmidrule(lr){2-5}
             & & \thead{Ball} & \thead{Shoe, IMU\\Head, Ball} & \thead{All}  \\
             \cmidrule(lr){2-5}\morecmidrules\cmidrule(lr){2-5}
             \parbox[t]{2mm}{\multirow{4}{*}{\rotatebox[origin=c]{90}{43\%}}} & BIP & - & 116.93 & \num{1.35e3} \\
             & PF & - & \cellcolor{gray!15} 0.61 & \cellcolor{gray!15} 0.61 \\
             & eBIP$-$ & - & \num{7.11e3} & \num{8.91e3} \\
             & eBIP & - & \cellcolor{green!15} 0.61 & \cellcolor{green!15} 0.59 \\
             \cmidrule(lr){2-5}
             \parbox[t]{2mm}{\multirow{4}{*}{\rotatebox[origin=c]{90}{82\%}}} & BIP & 6.67 & 205.04 & \num{2.74e3} \\
             & PF & 0.20 & \cellcolor{gray!15} 0.26 & \cellcolor{gray!15} 0.28 \\
             & eBIP$-$ & 5.38 & \num{8.44e3} & \num{9.46e3} \\
             & eBIP & \cellcolor{green!15} 0.13 & \cellcolor{green!15} 0.23 & \cellcolor{green!15} 0.24 \\
            \cmidrule(lr){2-5}\morecmidrules\cmidrule(lr){2-5}
        \end{tabular}
    \end{minipage}
    
    \caption{The left table indicates the mean squared error values for the first three joints of the robot at the time the ball is caught while the right table is the mean absolute error for the inferred ball position. A green box represents the best method and a gray box represents methods which are not statistically worse than the best method (Mann-Whitney U, $p < 0.05$). The values 43\% and 82\% indicate inference is performed after 43\% of the interaction is observed (corresponding to before the ball is thrown) and 82\% is observed (the ball is partway through its trajectory). The ball itself is not visible for the first 43\% as this is when it is occluded by the participant's hand. The standard error for eBIP is less than $\pm 0.01$ in all cases for the joint MSE and $\pm 0.015$ for the ball MAE.}
    \label{tab:big_state_results}
\end{table*}

\section{Experiments and Results}

We show the effectiveness of our proposed algorithm in the multimodal HRI scenario described in Sec.~\ref{sec:introduction} (Fig.~\ref{fig:teaser}). In this scenario, a human participant outfitted with a variety of sensors tosses a ball which is caught by a UR5~\cite{ur5} arm. The sensors can be broadly grouped into two categories: modalities that observe the human and modalities that observe the ball. Observations of the ball are unavailable while it is grasped by the human, due to occlusions, and do not become available until the ball is thrown. In empirical tests, we have observed that it is not possible for the robot to catch the ball using a purely reactive strategy given the limited time to react, kinematic constraints, phase lag, etc. Hence, we leverage the observation modalities of the human to predict how the robot should react while the human is still in the preparatory phase, i.e., the "wind up" for the throw. This strategy is fundamentally built upon a predictive approach -- we can begin reacting as early as possible and refine our predictions as more detailed observations become available.

\subsection{Experimental Setup}

The experiment is designed to emphasize the advantages of a multimodal observation set by having different sensors reveal different information about the true environment state at different points in time. However, throwing and catching are fast-paced actions requiring a high frequency observation rate and appropriately low computation times for inference; without these properties an HRI algorithm will likely fail at catching the ball in real experiments. We utilize sensor observations of 8 objects from 5 modalities: the positions of the human participant's hands and feet, inertial measurements of the throwing arm, pressure measurements from the soles of both left and right feet, the orientation of the head, the position of the ball being thrown, and the joint positions of the robot.
The observations were synchronized and collected at a frequency of $60$Hz.
The basis decomposition for each sensor object was chosen from a set of candidate basis spaces comprised of Polynomial, Gaussian, and Sigmoid functions -- standard choices in this type of application~\cite{bishop2006pattern} -- using the Bayesian Information Criterion, yielding a total state space dimension of $559$ dimensions.

During training, the ball was thrown from a distance of approx.~$3.7$m and was caught within a box grasped by the robot (the end effector used in this experiment actuates too slowly for in-hand interception). An initial set of $221$ demonstrations was provided via kinesthetic teaching in which the robot was manually operated by a human (top left of Fig.~\ref{fig:overview}) in order to catch the ball while joint positions were recorded.
These demonstrations provided the only source of prior knowledge for the interaction (for both state estimation and control); no inverse kinematics or other models were employed at any point in time.
We compare our algorithm, to the original BIP formulation, as well as particle filtering (PF). In all cases, the PF model used the same number of ensemble members as in eBIP and employed a systematic resampling scheme when the effective number of members was less than $E/2$.

\subsection{Results and Discussion}

The inference errors for the robot joints are shown in Table~\ref{tab:big_state_results}, along with the errors for the estimation of the location of the ball at the time of interception. For each data category, e.g., \{Shoe, IMU\}, only the indicated subset of sensor modalities is observed. We evaluate the prediction capabilities by observing a partial trajectory of sensor measurements and inferring the robot joint positions, as well as the position of the ball at the time of interception. We divide this into two categories: 43\% of the trajectory corresponds to the period of time in which the ball is in the human partner's hand and has yet to be thrown while 82\% corresponds to when the ball is still in the air and has yet to be caught.
The errors are listed in terms of the mean squared error for the first 3 joints of the robot (the wrist joints are less important in this scenario) and the mean absolute error of the ball prediction. Errors are computed via 10-fold cross validation over the randomly shuffled set of demonstrations, which also limits the maximum number of ensemble members used in both the eBIP and PF models to $198$.

\begin{figure*}[h]
    \centering
      \includegraphics[width=0.99\textwidth]{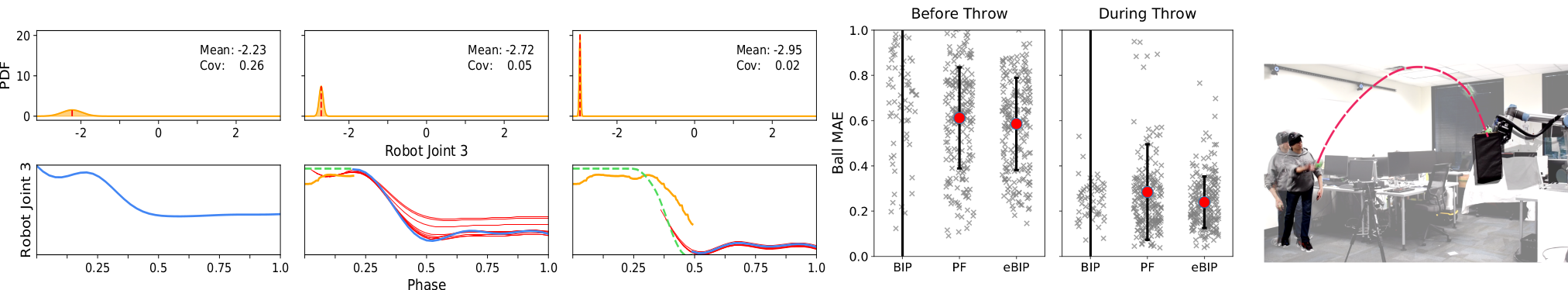}
    \caption{(\textbf{Left}) A sequence of frames from different time points during an interaction. Top: the PDF of the third robot joint; the initial uncertainty is high and decreases over time. Bottom: the inferred trajectory for the robot joint. The blue line indicates the current prediction while the red lines indicate the predictions for the past 10 time steps. The yellow line is the actual response from the robot while it attempts to follow the inferred trajectory, and the dashed green line indicates the expected trajectory from the demonstration. (\textbf{Center}) The ball MAE of the \{All\} subset. While overall error decreases for both PF and eBIP over time, only eBIP experiences a reduced variance. (\textbf{Right}) A blindfolded user throws a ball which is, in turn, caught by the robot.}
    \label{fig:uncertainty}
\end{figure*}

\textbf{Prior Approximation Errors: }
Results  in  Table  I  show that attempting to model the demonstrations with a parametric Gaussian model yields a poor approximation and leads to an incorrect estimate of the initial uncertainty. This is supported by the fact that both BIP (Gaussian prior) and eBIP$^-$ (mixture model  prior)  produce  predictions  that  are  many  orders  of magnitude  from  the  true  state.  In  the case of eBIP$^-$,  expectation maximization regularly produced non-positive semi-definite covariance matrices (using 1 component as determined by BIC), indicating a poor fit to the data set. As a result, we were forced to use the sample mean and covariance of the demonstrations for the Gaussian prior as  in  BIP,  from  which  the  initial  ensemble  is sampled.  The PF  and  eBIP  methods,  on  the  other  hand,  were  initialized directly from the demonstrations without making an assumption about the parametric family of the true (unknown) distribution. As a result both methods fared much better, however, eBIP significantly  more  so  as  it  achieved  the  best  result  in  every category (see green box).

\textbf{Linearization Errors:}
We can also observe that BIP certainly suffers from linearization errors.
Since eBIP$^-$ models the prior distribution with the same unimodal Gaussian as BIP, we expect it to suffer from the same prior approximation errors. However, we see from Table~\ref{tab:big_state_results} that BIP yields a \num{1.04e15} joint prediction MSE error when 82\% of the ball trajectory is observed while eBIP$^-$ only yields a joint prediction MSE of 10.52. The remainder of this error is due to the different update methods and linearization errors inherent to BIP.
\begin{figure}[ht!]
    \centering
    \includegraphics[width=0.95\columnwidth, trim={3cm 0 4cm 0}, clip]{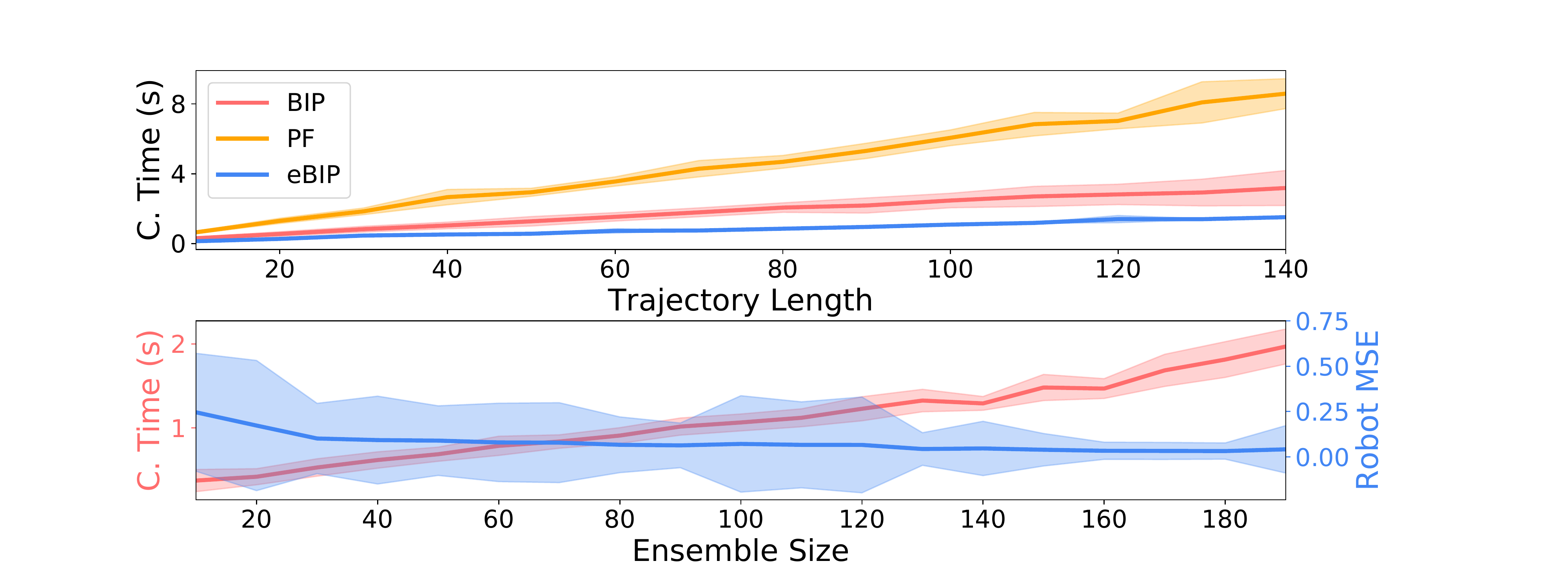}
    \caption{Top: computation time required for filtering observations of varying lengths. Bottom: the computation time-accuracy-ensemble size trade-off for eBIP.}
    \label{fig:comp_time}
\end{figure}

\textbf{Errors Resulting from Increased State Dimension: }
These results also show that both the PF and eBIP$^-$ produce worse inference results as the number of active modalities increases. For example, the PF predictions result in a joint MSE of $0.11$ when only the trajectory of the ball was observed, but a MSE of $0.26$ when all modalities were observed. We can rule out both prior approximation error and linearization errors, since PF utilizes the demonstrations and nonlinear system functions directly as in eBIP. Therefore, we conclude that the number of ensemble members is simply too low to provide accurate coverage of the state space leading to sample degeneracy as a result of importance sampling. In the case of eBIP$^-$, the increasing error is due to the approximation errors stemming from the prior distribution, since otherwise the algorithm is identical to eBIP.

\textbf{Errors Resulting from Additional Modalities: }
Lastly, we observe that the introduction of additional modalities does not always yield an increase in inference accuracy, although it may provide other benefits depending on the modalities. This is evident when comparing the joint MSE prediction errors of eBIP for the \{Shoe, IMU\} data set and that of the \{Shoe, IMU, Head\} data set. The introduction of the head modality actually \emph{increases} the inference error when 43\% of the trajectory is observed, which is when the head modality is most relevant. This becomes particularly evident when looking at the MAE results of the ball; adding additional sensor modalities increases the inference error of the final ball position by a factor of 2. However, we also gain the ability to initially predict the ball position much earlier in the interaction, before the ball is visible.
This process is visualized in Fig.~\ref{fig:uncertainty} through the uncertainty in the inferred joint trajectories.
The width of the catchable region is approximately 180cm. Hence, the fact that we can predict the interception point to within about 60cm, or $1/3$ of our operating region, \emph{before the ball is visible} is quite significant. By the time the ball is in the air (82\% of the trajectory is observed), we have further narrowed the prediction down to 23cm with additional modalities. While this error is still significantly higher than the 13cm error produced by incorporating only the ball, we observed empirically that a 23cm error still results in a catch in most cases and justifies the inclusion of additional modalities. Given that the radius of the ball itself is 4.5cm and the width of the box is 32cm, this amount of error is adequate and is offset by the proactive behavior of the robot in this setting. Still, the above results suggests that there may be substantial benefit to the ability of the inference process to switch modalities on and off in real-time according to the context.

\textbf{Real-time interactions:}
Results of the real-time interaction and reproduction with the robot can be seen in Fig.~\ref{fig:throw_sequence}. To minimize the computation time required for inference, the number of ensemble members was limited to $80$. This setup ensured a reasonable trade-off between accuracy and computational performance as shown in Fig.~\ref{fig:comp_time}. Three observation can be made from the image sequences in Fig.~\ref{fig:throw_sequence}: the throws all have significantly different trajectories (the top throw has a low apex and fast velocity while the bottom throw has a high apex and low velocity), the robot is already moving into position before the ball is thrown (second image in each sequence), and the robot catches the ball in a different pose for each throw (last image in each sequence). Recordings of a variety of throwing experiments can be found in the accompanying video. To avoid habituation or any unconscious effort to throw the ball directly at the robot, we also performed a set of experiments in which the user threw the ball while in a blindfolded condition, see Fig.~\ref{fig:uncertainty} (right) for an example. Even under this condition, the ball was successfully caught $12$ out of $20$ times, for a success rate of $60$\%. We noticed, however, that in this condition the user frequently threw the ball outside the robot's reach.

\section{Conclusions}
In this paper, we introduced ensemble Bayesian Interaction Primitives and discussed their application to state estimation, inference, and control in challenging, fast-paced HRI tasks with many data sources. We discussed an ensemble-based approach to Bayesian inference in eBIP, which requires neither an explicit formation of a covariance matrix, nor a measurement model, resulting in significant computational speed-ups. The approach allows for fast inference from high-dimensional, probabilistic models and avoids typical sources for inaccuracies, e.g., linearization and Gaussian priors. In our real-robot experiments, a relatively small number of ensemble members produced a reasonable trade-off between accuracy and computational performance. However, our results also indicate that the uncontrolled inclusion of many data sources is not always beneficial. Some modalities may introduce spurious correlations or significant amounts of noise into the filtering process, thereby harming the accuracy of predictions. These challenges may be overcome by incorporating feature selection mechanisms, or by switching individual modalities on and off according to context.

\section*{Acknowledgments}

This work was supported by a grant from the Honda Research Institute and by the National Science Foundation under Grant No.~(IIS-1749783).

\bibliographystyle{plainnat}
\bibliography{references}

\begin{thebibliography}{28}
\providecommand{\natexlab}[1]{#1}
\providecommand{\url}[1]{\texttt{#1}}
\expandafter\ifx\csname urlstyle\endcsname\relax
  \providecommand{\doi}[1]{doi: #1}\else
  \providecommand{\doi}{doi: \begingroup \urlstyle{rm}\Url}\fi

\bibitem[Amor et~al.(2014)Amor, Neumann, Kamthe, Kroemer, and
  Peters]{amor2014interaction}
Heni~Ben Amor, Gerhard Neumann, Sanket Kamthe, Oliver Kroemer, and Jan Peters.
\newblock Interaction primitives for human-robot cooperation tasks.
\newblock In \emph{Robotics and Automation (ICRA), 2014 IEEE International
  Conference on}, pages 2831--2837. IEEE, 2014.

\bibitem[Bishop(2006)]{bishop2006pattern}
Christopher~M Bishop.
\newblock Pattern recognition and machine learning.
\newblock 2006.

\bibitem[Burgers et~al.(1998)Burgers, Jan~van Leeuwen, and
  Evensen]{burgers1998analysis}
Gerrit Burgers, Peter Jan~van Leeuwen, and Geir Evensen.
\newblock Analysis scheme in the ensemble kalman filter.
\newblock \emph{Monthly weather review}, 126\penalty0 (6):\penalty0 1719--1724,
  1998.

\bibitem[Calinon and Billard(2008)]{Calinon08ICRA}
Sylvain Calinon and Aude Billard.
\newblock A framework integrating statistical and social cues to teach a
  humanoid robot new skills.
\newblock In \emph{Proc. {IEEE} Intl Conf. on Robotics and Automation ({ICRA}),
  Workshop on Social Interaction with Intelligent Indoor Robots}, May 2008.

\bibitem[Campbell and Amor(2017)]{campbell2017bayesian}
Joseph Campbell and Heni~Ben Amor.
\newblock Bayesian interaction primitives: A slam approach to human-robot
  interaction.
\newblock In \emph{Conference on Robot Learning}, pages 379--387, 2017.

\bibitem[Chen et~al.(2017)Chen, Wu, Duan, Guan, and Rojas]{chen2017}
L.~Chen, H.~Wu, S.~Duan, Y.~Guan, and J.~Rojas.
\newblock Learning human-robot collaboration insights through the integration
  of muscle activity in interaction motion models.
\newblock In \emph{2017 IEEE-RAS 17th International Conference on Humanoid
  Robotics (Humanoids)}, pages 491--496, Nov 2017.

\bibitem[Coen(2001)]{Coen:2001}
Michael~H. Coen.
\newblock Multimodal integration: A biological view.
\newblock In \emph{Proceedings of the 17th International Joint Conference on
  Artificial Intelligence - Volume 2}, IJCAI'01, pages 1417--1424, San
  Francisco, CA, USA, 2001. Morgan Kaufmann Publishers Inc.

\bibitem[Cui et~al.(2018)Cui, Poon, Miro, Yamazaki, Sugimoto, and
  Matsubara]{Cui2018}
Yunduan Cui, James Poon, Jaime~Valls Miro, Kimitoshi Yamazaki, Kenji Sugimoto,
  and Takamitsu Matsubara.
\newblock Environment-adaptive interaction primitives through visual context
  for human--robot motor skill learning.
\newblock \emph{Autonomous Robots}, Aug 2018.

\bibitem[Dermy et~al.(2019)Dermy, Charpillet, and Ivaldi]{dermy2019multi}
Oriane Dermy, Fran{\c{c}}ois Charpillet, and Serena Ivaldi.
\newblock Multi-modal intention prediction with probabilistic movement
  primitives.
\newblock In \emph{Human Friendly Robotics}, pages 181--196. Springer, 2019.

\bibitem[Evensen(2003)]{evensen2003ensemble}
Geir Evensen.
\newblock The ensemble kalman filter: Theoretical formulation and practical
  implementation.
\newblock \emph{Ocean dynamics}, 53\penalty0 (4):\penalty0 343--367, 2003.

\bibitem[Ewerton et~al.(2015)Ewerton, Neumann, Lioutikov, Amor, Peters, and
  Maeda]{ewerton2015}
Marco Ewerton, Gerhard Neumann, Rudolf Lioutikov, Heni~Ben Amor, Jan Peters,
  and Guilherme Maeda.
\newblock Learning multiple collaborative tasks with a mixture of interaction
  primitives.
\newblock In \emph{2015 IEEE International Conference on Robotics and
  Automation (ICRA)}, pages 1535--1542, May 2015.
\newblock \doi{10.1109/ICRA.2015.7139393}.

\bibitem[Haseltine and Rawlings(2005)]{haseltine2005critical}
Eric~L Haseltine and James~B Rawlings.
\newblock Critical evaluation of extended kalman filtering and moving-horizon
  estimation.
\newblock \emph{Industrial \& engineering chemistry research}, 44\penalty0
  (8):\penalty0 2451--2460, 2005.

\bibitem[Iqbal and Riek(2019)]{iqbal2019human}
Tariq Iqbal and Laurel~D Riek.
\newblock Human-robot teaming: Approaches from joint action and dynamical
  systems.
\newblock \emph{Humanoid Robotics: A Reference}, pages 2293--2312, 2019.

\bibitem[Lawson and Hansen(2004)]{lawson2004implications}
W~Gregory Lawson and James~A Hansen.
\newblock Implications of stochastic and deterministic filters as
  ensemble-based data assimilation methods in varying regimes of error growth.
\newblock \emph{Monthly weather review}, 132\penalty0 (8):\penalty0 1966--1981,
  2004.

\bibitem[Lee and Nakamura(2010)]{donghuilee2010}
Dongheui Lee and Yoshihiko Nakamura.
\newblock Mimesis model from partial observations for a humanoid robot.
\newblock \emph{The International Journal of Robotics Research}, 29\penalty0
  (1):\penalty0 60--80, 2010.

\bibitem[Lei et~al.(2010)Lei, Bickel, and Snyder]{lei2010comparison}
Jing Lei, Peter Bickel, and Chris Snyder.
\newblock Comparison of ensemble kalman filters under non-gaussianity.
\newblock \emph{Monthly Weather Review}, 138\penalty0 (4):\penalty0 1293--1306,
  2010.

\bibitem[Lundquist et~al.(2015)Lundquist, Sjanic, and
  Gustafsson]{Lundquist:2015}
Christian Lundquist, Zoran Sjanic, and Fredrik Gustafsson.
\newblock \emph{Statistical Sensor Fusion: Exercises}.
\newblock Studentlitteratur AB, Sweden, 2015.

\bibitem[Mandel(2006)]{mandel2006efficient}
Jan Mandel.
\newblock \emph{Efficient implementation of the ensemble Kalman filter}.
\newblock University of Colorado at Denver and Health Sciences Center, Center
  for Computational Mathematics, 2006.

\bibitem[Miller et~al.(1994)Miller, Ghil, and Gauthiez]{miller1994advanced}
Robert~N Miller, Michael Ghil, and Francois Gauthiez.
\newblock Advanced data assimilation in strongly nonlinear dynamical systems.
\newblock \emph{Journal of the atmospheric sciences}, 51\penalty0 (8):\penalty0
  1037--1056, 1994.

\bibitem[Noda et~al.(2014)Noda, Arie, Suga, and Ogata]{noda2014multimodal}
Kuniaki Noda, Hiroaki Arie, Yuki Suga, and Tetsuya Ogata.
\newblock Multimodal integration learning of robot behavior using deep neural
  networks.
\newblock \emph{Robotics and Autonomous Systems}, 62\penalty0 (6):\penalty0
  721--736, 2014.

\bibitem[Oguz et~al.(2018)Oguz, Zhou, and Wollherr]{ogur2018}
Ozgur~S. Oguz, Zhehua Zhou, and Dirk Wollherr.
\newblock A hybrid framework for understanding and predicting human reaching
  motions.
\newblock \emph{Frontiers in Robotics and AI}, 5:\penalty0 27, 2018.

\bibitem[Piaget(1955)]{piaget1955child}
Jean Piaget.
\newblock \emph{The Child's Construction of Reality}.
\newblock Routledge \& Paul, 1955.

\bibitem[Rabiner(1989)]{rabiner1989}
L.~R. Rabiner.
\newblock A tutorial on hidden markov models and selected applications in
  speech recognition.
\newblock \emph{Proceedings of the IEEE}, 77\penalty0 (2):\penalty0 257--286,
  Feb 1989.

\bibitem[Rao et~al.(2017)Rao, De~Deuge, Nourani-Vatani, Williams, and
  Pizarro]{rao2017multimodal}
Dushyant Rao, Mark De~Deuge, Navid Nourani-Vatani, Stefan~B Williams, and Oscar
  Pizarro.
\newblock Multimodal learning and inference from visual and remotely sensed
  data.
\newblock \emph{The International Journal of Robotics Research}, 36\penalty0
  (1):\penalty0 24--43, 2017.

\bibitem[Robots()]{ur5}
Universal Robots.
\newblock {UR5 Robot Arm}.
\newblock \url{https://www.universal-robots.com/products/ur5-robot/}.
\newblock [Online; accessed May 15, 2019].

\bibitem[Rozo et~al.(2016)Rozo, Silv\'erio, Calinon, and
  Caldwell]{Rozo16Frontiers}
Leonel Rozo, Joao Silv\'erio, Sylvain Calinon, and Darwin~G. Caldwell.
\newblock Learning controllers for reactive and proactive behaviors in
  human-robot collaboration.
\newblock \emph{Frontiers in Robotics and {AI}}, 3\penalty0 (30):\penalty0
  1--11, June 2016.
\newblock Specialty Section Robotic Control Systems.

\bibitem[Thomaz et~al.(2016)Thomaz, Hoffman, Cakmak,
  et~al.]{thomaz2016computational}
Andrea Thomaz, Guy Hoffman, Maya Cakmak, et~al.
\newblock Computational human-robot interaction.
\newblock \emph{Foundations and Trends{\textregistered} in Robotics},
  4\penalty0 (2-3):\penalty0 105--223, 2016.

\bibitem[Thrun et~al.(2005)Thrun, Burgard, and Fox]{thrun2005probabilistic}
Sebastian Thrun, Wolfram Burgard, and Dieter Fox.
\newblock \emph{Probabilistic robotics}.
\newblock MIT press, 2005.

\end{thebibliography}

\end{document}